\pdfoutput=1

\documentclass[11pt]{article}

\usepackage[]{acl}

\usepackage{times}
\usepackage{latexsym}
\usepackage{graphicx}
\usepackage{amsmath}

\usepackage[T1]{fontenc}

\usepackage[utf8]{inputenc}

\usepackage{microtype}

\usepackage{inconsolata}

\usepackage{booktabs}  
\usepackage{multirow}
\usepackage{makecell}  

\newcommand{\datasetname}[0]{PEFT-U}

\title{\datasetname: Parameter-Efficient Fine-Tuning for User Personalization}

\author{
  Christopher Clarke\hspace{10pt} Yuzhao Heng\hspace{10pt}  \textbf{Lingjia Tang\hspace{10pt}  Jason Mars\hspace{10pt}}    \vspace{0.3cm}\\
  \text{Computer Science \& Engineering} \\
    \text{University of Michigan} \\ \text{Ann Arbor, MI}\\
    \text{\{csclarke, stefanhg, lingjia, profmars\}@umich.edu} \\
}

\begin{document}
\maketitle
\begin{abstract}
The recent emergence of Large Language Models (LLMs) has heralded a new era of human-AI interaction. These sophisticated models, exemplified by Chat-GPT and its successors, have exhibited remarkable capabilities in language understanding. However, as these LLMs have undergone exponential growth, a crucial dimension that remains understudied is the personalization of these models. Large foundation models such as GPT-3 etc. focus on creating a universal model that serves a broad range of tasks and users. This approach emphasizes the model's generalization capabilities, treating users as a collective rather than as distinct individuals. While practical for many common applications, this one-size-fits-all approach often fails to address the rich tapestry of human diversity and individual needs. To explore this issue we introduce the \textbf{\datasetname{} Benchmark}: a new dataset for building and evaluating NLP models for user personalization. \datasetname{} consists of a series of user-centered tasks containing diverse and individualized expressions where the preferences of users can potentially differ for the same input. Using \datasetname{}, we explore the challenge of efficiently personalizing LLMs to accommodate user-specific preferences in the context of diverse user-centered tasks.
\end{abstract}

\section{Introduction}

Large Language Models (LLMs) have shown tremendous capability in performing complex tasks such as reasoning, summarization, creative writing, etc. Through the scaling of these models, both in size ($>$ 1B parameters) and data ($>$ 1 Trillion tokens) these models have achieved remarkable performance on a wide range of natural language understanding and generation tasks \cite{touvron2023llama, henighan2020scaling}. However, even as the generalization capability of LLMs has grown, one crucial dimension that has been understudied is the personalization of these models \cite{salemi2023lamp, human-centered-23}.

At its core, personalization is about tailoring AI-driven interactions to the individual preferences, needs, and idiosyncrasies of each user \cite{salemi2023lamp, welch-etal-2022-leveraging, clarke-etal-2023-rule, kang2022personalized}. In many real-world scenarios, users have unique preferences, contexts, and expectations, which are currently incapable of being effectively accommodated by the generalized LLMs available today. These traditional LLMs have predominantly adhered to a "one-size-fits-all" approach \cite{touvron2023llama,clarke-etal-2022-one, openai2023gpt4, bubeck2023sparks}, offering a single, uniform model capable of serving all users and tasks alike. While this approach is undoubtedly valuable in many scenarios, it falls short when it comes to accommodating the rich tapestry of human diversity where people are not uniform, and their linguistic and communicative preferences vary widely \cite{li2023teach, zhang-etal-2018-personalizing}. 

Existing works in NLP have highlighted the need for user perspective in language modeling particularly when dealing with intrinsically subjective applications such as Hate Speech Detection and Sentiment Analysis where differing perspectives are common \cite{davani-etal-2022-dealing, sang2021origin, geva-etal-2019-modeling, kanclerz-etal-2022-ground, welch-etal-2022-leveraging}. Research has shown that accounting for user perspective and personalization is essential for building robust and effective models \cite{davani-etal-2022-dealing, sang2021origin, geva-etal-2019-modeling, kanclerz-etal-2022-ground, welch-etal-2022-leveraging}. However, despite this glaring need, existing resources fail to model and cater to these differing perspectives. When curated, NLP resources tend to have an intrinsic bias towards the majority perspective due to their reliance on voting for ground truth. As such they fail to adequately represent diverse user preferences and individualized expressions, further contributing to a lack of personalization \cite{davani-etal-2022-dealing, sang2021origin, geva-etal-2019-modeling, kanclerz-etal-2022-ground}.

To combat these challenges we introduce the \textbf{\datasetname{} Benchmark}. \datasetname{} consists of over 13+ personalized tasks and 15k+ users across domains such as Hate Speech, Sentiment/Emotion, and Humor. In contrast to other resources, the \datasetname{} benchmark uniquely tests complex scenarios where LLMs are faced with differing user perspectives even when facing the same input. To the best of our knowledge, this benchmark is the first of its kind to focus on modeling user preferences in NLP with an emphasis on identical inputs that require different model outputs depending upon the user. Using \datasetname{} we explore a range of strategies for efficiently compartmentalizing user perspectives. In particular, we implement and empirically analyze a series of personalized prompting approaches (non-parametric) vs tuning and compartmentalizing user-level knowledge (parametric) for personalized tasks showing that personalized models are crucial to providing users with more accurate results representative of their actual perspectives. We publicly release our code, models, and benchmark\footnote{\url{https://github.com/ChrisIsKing/Parameter-Efficient-Personalization}}.

\section{\datasetname{} Benchmark}
The \datasetname{} benchmark aims to assess the efficacy of language models in producing personalized outputs based on user-specific information. 

\paragraph{Data Collection} To generate high-quality data samples representative of differing user perspectives we focus on curating subjective problems where the model is forced to respect the user's points of view e.g. Hate Speech Detection. Typically NLP resources for these problem areas employ an annotation process where correctness is determined via majority vote and outliers are discarded. This often results in the overlooking of the subtleties of the user's perspective, ultimately leading to potential group biases and inaccuracies in the data. In contrast, we reconstruct these data sources framing individual annotators as distinct users to capture these important nuances. To avoid the possible influence of noisy/bad annotators we take into account their contribution level to the annotation process and discard low-quality users. Additionally, we discard users with less than $n=10$ samples in their training and test sets respectively. As shown in table \ref{tab:benchmark} \datasetname{} consists of 13+ personalized tasks and 15k+ users with each task gaining a maximum Krippendorff's alpha ($\alpha$) of 0.5 \cite{Krippendorff2011ComputingKA} across the domains of Hate/Abuse, Humor, and Emotion/Sentiment as shown in table \ref{tab:benchmark}. For each dataset, we construct a unique instruction-style prompt to guide the LLM to generate the desired output. More details on each of the specific datasets, our reconstruction process, and their prompts are provided in Appendix \ref{apd:datasets}.






\paragraph{User Disagreement}
As shown in table \ref{tab:benchmark}, we enforce that all personalized tasks must obtain a Krippendorff's alpha score of $(\alpha \leq 0.5)$. This requirement is created to assess the ability to capture differing user perspectives even when facing the same input. Krippendorff's alpha $(\alpha)$ is a reliability coefficient developed to measure the agreement among annotators. When used in data curation, data is usually considered reliable when $(\alpha \geq 0.8)$. By enforcing all datasets to have low agreement scores, we force the model to rely on respective user information to generate its output.

\begin{table*}[]
    \tiny
    \resizebox{\textwidth}{!}{%
    \begin{tabular}{ l l c c c c}

        \toprule
        \multicolumn{2}{c}{\thead{Dataset}} & \multirow{2}{*}[-0.25em]{\thead{\# Users}} & \multirow{2}{*}[-0.25em]{\thead{Avg \# examples\\per user}} & \multirow{2}{*}[-0.25em]{\thead{Avg \# users\\per text}} & \multirow{2}{*}[-0.25em]{\thead{Krippendorff's\\Alpha}} \\
        \addlinespace[0.25em] \cline{1-2} \addlinespace[0.5em] 
        \thead{Domain} & \thead{Name} & \\

        \midrule
        \multirow{6}{*}{Hate+Abuse} 
            & HateXplain \cite{mathew2022hatexplain} & 253  & 238.9  & 3.0  & 0.46 \\
            & GabHate \cite{Kennedy2018TheGH} & 18 & 4807.16  & 3.12  & 0.24 \\
            & MeasuringHateSpeech \cite{sachdeva-etal-2022-measuring} & 7912  & 17.13  & 3.42  & 0.47 \\
            & TweetEval \cite{röttger2022contrasting} & 20 & 200.0 & 200.0 & 0.14 \\
            & UnhealthyConversations \cite{price-etal-2020-six} & 588 & 386.77 & 4.64 & 0.28 \\
            & WikiDetox Aggression \cite{wulczyn2017ex} & 4053 & 336.84 & 11.78 & 0.43 \\

        \midrule
        \multirow{3}{*}{Sentiment} 
            & GoEmotion \cite{demszky2020goemotions} & 82 & 1859.95 & 2.83 & 0.44 \\
            & StudEmo \cite{ngo-etal-2022-studemo} & 25 & 296.44 & 1.43 & 0.18 \\
            & Subjective Discourse \cite{ferracane-etal-2021-answer}$^\ast$ & 68 & 9.26 & 6.20 & 0.50/0.01/0.01 \\

        \midrule
        \multirow{2}{*}{Humor} 
            & Cockamamie \cite{pmlr-v97-gultchin19a} & 1878 & 489.33 & 7.65 & 0.08 \\
            & EPIC \cite{frenda-etal-2023-epic} & 74 & 191.51 & 4.72 & 0.19 \\
        \bottomrule
    \end{tabular}}
    \caption{\datasetname{} Benchmark: We design a large-scale benchmark for personalized model training and evaluation consisting of 13+ personalized tasks across 15k+ users with each task obtaining a Krippendorff's alpha ($\alpha$) $<$ 0.5 per task. $^\protect\ast$Subjective Discourse consists of 3 different sentiment tasks.} 
    \label{tab:benchmark}
\end{table*}

\section{Modularity + Personalization}
When exploring the problem of personalization, one possible solution would be the allocation of a dedicated LLM for each user. However, deploying a separate personalized model for each user would incur significant compute costs in production. In addition, the balance between embedding generalized and personalized knowledge in these models remains unclear. Thus providing such a solution is prohibitive in this era of large language models. Recent works in Modular Deep Learning \cite{liu2022fewshot, pfeiffer2023modular, hu2021lora, houlsby2019parameterefficient}, seek to optimize and further tune these LLMs without having to update all the model parameters. These methods typically introduce a small number of additional parameters and update these parameters in a model while freezing the remaining weights, thus limiting the computational resources required. This is often done to enable multi-task learning or to introduce new updates in the training data without the need for training from scratch. 

In our experiment setting, we shift this paradigm from multi-task learning to multi-user learning focusing on the research question of \textit{"How to efficiently personalize large language models for subjective tasks?"}. As such, we empirically analyze personalized prompting approaches (non-parametric) vs efficiently tuning and compartmentalizing user-level knowledge (parametric) for personalized tasks.

\begin{table*}[]
    
    \resizebox{\textwidth}{!}{%
    \begin{tabular}{@{\extracolsep{2pt}} l c c  c c c c c c c c c c c c c  c @{}}

        \toprule
        \multirow{3}{*}[-0.85em]{\thead{Method}} & \multirow{3}{*}[-0.85em]{\thead{Size}} & \multirow{3}{*}[-0.85em]{\thead{\#Params}} & \multicolumn{13}{c}{\thead{Dataset}} & \multirow{3}{*}[-0.85em]{\thead{Average}} \\
        
        \addlinespace[0.125em] \cline{4-16} \addlinespace[0.125em]
        & & & \multicolumn{6}{c}{\thead{Hate+Abuse}} & \multicolumn{2}{c}{\thead{Sentiment}} & \multicolumn{2}{c}{\thead{Humor}} & \multicolumn{3}{c}{\thead{Sentiment}}\\
        
        \addlinespace[0.125em] \cline{4-9} \cline{10-11} \cline{12-13} \cline{14-16} \addlinespace[0.125em]
        & & & \thead{Hate\\Xplain} & \thead{Gab\\Hate} & \thead{MHS} & \thead{Tweet\\Eval} & 
        \thead{UHC} & \thead{Wiki\\Detox} & \thead{Go\\Emotion} & \thead{Stud\\Emo} & \thead{Cockamamie} & \thead{EPIC} & \thead{SD(D)} & \thead{SD(QS)} & \thead{SD(RS)} & \\
        
        \midrule
        \textbf{Zero/Few-Shot (k=3)} & - & 0 & 47.4 & 81.9 & 26.1 & 61.5 & 55.2 & 61.4 & 53.3 & 27.2 & 97.1 & 63.3 & 16.0 & 20.5 & 11.3 & 47.9 \\

        \midrule
        \textbf{LoRA} & 3.8M & 880K & 54.9 & 88.6 & 27.4 & 69.5 & 73.4 & 81.0 & 66.3 & 67.7 & 97.2 & 67.9 & 29.4 & 30.7 & 18.9 & 59.5 \\
        
        \textbf{Prefix Tuning} & 1.5M & 370K & 48.8 & 85.4 & \textbf{45.5} & 61.8 & 66.0 & 72.7 & 60.1 & 32.9 & 97.1 & 66.7 & 7.7 & 11.7 & 8.1 & 51.1 \\
        
        \textbf{P-Tuning} & 128K & 1.8M & 48.7 & 81.9 & 29.2 & 61.3 & 56.8 & 62.3 & 53.8 & 27.6 & 97.0 & 62.8 & 16.6 & 19.6 & 11.1 & 48.4 \\
        
        \textbf{Prompt Tuning} & 128K & 31K & 49.3 & 82.8 & 28.7 & 59.5 & 56.3 & 63.3 & 52.1 & 27.6 & 97.1 & 61.1 & 10.5 & 20.5 & 9.9 & 47.6 \\
        
        \textbf{Adapters} & 14M & 3.5M & \textbf{59.0} & \textbf{89.1} & 38.9 & \textbf{70.5} & \textbf{77.5} & \textbf{84.0} & \textbf{68.4} & \textbf{83.7} & \textbf{97.2} & \textbf{68.5} & \textbf{35.3} & \textbf{39.0} & \textbf{25.6} & \textbf{64.4} \\
        
        \textbf{IA\textasciicircum{}3} & 450K & 111K & 48.6 & 86.7 & 26.5 & 62.3 & 61.6 & 70.0 & 58.9 & 27.6 & 97.1 & 64.3 & 19.0 & 24.1 & 13.5 & 50.8 \\

        \bottomrule
    \end{tabular}}
    \caption{Results of PEFT Methods on the \datasetname{} Benchmark: This table shows the macro accuracy of each PEFT method in comparison to Zero/Few-shot prompting on Flan-T5 \cite{flan-t5}.}
    \label{tab:results}
\end{table*}

\section{Benchmark Evaluation}
To quantify the challenge the \datasetname{} benchmark presents, we evaluate the performance of a range of parameter-efficient methods compared to zero/few-shot prompting approaches.

\paragraph{Methods} We implement and evaluate 7 different parameter-efficient methods for personalizing LLMs using Flan-T5 \cite{flan-t5}. These methods consist of: 

1) \textbf{Zero-shot/Few-shot Prompting}: Using $k=3$ random samples of user data we construct an instruction-style prompt for inference. 

2) \textbf{LoRa} \cite{hu2021lora}: injects trainable rank decomposition matrices into each layer of the Transformer architecture.

3) \textbf{Adapters} \cite{houlsby2019parameterefficient} add a trainable bottleneck layer after the feedforward network in each Transformer layer. 

4) \textbf{Prompt Tuning} \cite{lester2021power} introduces an additional $k$ tunable tokens per downstream task prepended to the input text and trained end-to-end.

5) \textbf{Prefix-Tuning} \cite{li-liang-2021-prefix} prepends task-specific trainable vectors to the input of multi-head attention in each Transformer layer that is attended to as virtual tokens.

6) \textbf{P-Tuning} \cite{liu-etal-2022-p-tuning} employs trainable continuous prompt embeddings in concatenation with discrete prompts. 

7) \textbf{IA\^{}3} \cite{liu2022fewshot} introduces trainable vectors $l_w$ into different components of the transformer which perform element-wise rescaling of inner model activations.

    






\paragraph{Training} \label{train}
We train all models with AdamW \citep{adamw} and a weight decay of 0.01 on NVIDIA RTX 3090 24GB GPUs. We use a learning rate of 2e-5, batch size of 16, and a linear learning rate warmup over the first 10\% steps with a cosine schedule for 8 epochs over multiple runs with varied random seeds.

\section{Results}

\paragraph{Evaluation Metrics} We consider two performance metrics: (1) average per-user accuracy per task and (2) average accuracy for all tasks.

\subsection{Few/Zero-shot vs PEFT}
Table \ref{tab:results} shows our results analyzing existing PEFT methods in comparison to few/zero-shot prompting techniques. From these results, we show that personalizing models is crucial to providing users with more accurate results representative of their actual perspectives. Notably, zero/few-shot prompting falls short of adequately representing user viewpoints compared to its trained counterparts being outperformed on average by all methods except for Prompt Tuning. Across all methods, results show that Adapters performs the best outperforming on 12 out of the 13 \datasetname{} tasks and achieving an overall accuracy score of 64.4\% compared to 59.5\% of LoRa in 2nd. The presented results underscore the complexity of the \datasetname{} benchmark, revealing the challenges inherent in achieving consistently high performance across diverse tasks and datasets. While personalized fine-tuning methods exhibit superior accuracy compared to traditional few/zero-shot prompting techniques, the variations in performance among different PEFT methods as well as the performance on datasets such as Subjective Discourse and MeasuringHateSpeech indicate that the benchmark presents a multifaceted challenge. The nuances of user personalization, model size, and parameter tuning significantly impact the effectiveness of these methods. This observed diversity in performance across methods suggests that there is no one-size-fits-all solution, and further investigation is imperative.

\begin{table}[]
    \resizebox{\columnwidth}{!}{%
    \begin{tabular}{l c c c c}
        
        \toprule
        \thead{Method} & \thead{Original\\Params} & \thead{Adjusted\\Params} & \thead{Original\\Acc} & \thead{Acc\\w/ Reduced Params} \\

        \midrule
        \textbf{LoRA} & 880K & 111K & \textit{69.5} & \textbf{66.2}  \\
        \textbf{Prefix Tuning} & 370K & 111K & \textit{61.8} & 61.5 \\
        \textbf{P-Tuning} & 1.6M & 111K & \textit{61.7} & 61.7 \\
        \textbf{Prompt Tuning} & 15K & 111K & \textit{59.8} & 50.0 \\
        \textbf{Adapters} & 3.5M & 111K & \textit{70.5} & 64.2 \\
        \textbf{IA\textasciicircum{}3} & 111K & - & \textit{62.3} & \textbf{-} \\ 
        
        \bottomrule
    \end{tabular}}
    \caption{Results on TweetEval task for \datasetname{} with equal number of trainable parameters.}
    \label{tab:results-2}
\end{table}

\subsection{Impact of Number of Parameters}
Given the performance of Adapters, we sought to understand whether its performance was due to the number of trainable parameters. As such, we systematically varied the parameter count across each method on the TweetEval Task. Notably, we observed nuanced patterns across different PEFT methods. As shown in table \ref{tab:results-2}, with reduced parameters all methods except for P-tuning suffered a decrease in overall performance. However, LoRa with equal trainable parameters was able to outperform Adapters.

\section{Related Works}
Prior works have highlighted the need for user perspective particularly when dealing with intrinsically subjective applications where differing perspectives are common \cite{davani-etal-2022-dealing, sang2021origin, geva-etal-2019-modeling, kanclerz-etal-2022-ground, welch-etal-2022-leveraging}. Research has shown that accounting for user perspective and personalization is essential for building robust and effective models \cite{davani-etal-2022-dealing, sang2021origin, geva-etal-2019-modeling, kanclerz-etal-2022-ground, welch-etal-2022-leveraging}. However, despite this glaring need, existing resources fail to model and cater to these differing perspectives. When curated, NLP resources tend to have an intrinsic bias towards the majority perspective due to their reliance on voting for ground truth. As such they fail to adequately represent diverse user preferences and individualized expressions, further contributing to a lack of personalization \cite{davani-etal-2022-dealing, sang2021origin, geva-etal-2019-modeling, kanclerz-etal-2022-ground}. Other benchmarks such as \citet{salemi2023lamp} highlight the importance of personalization LLMs, however, \datasetname{} uniquely factors cases of conflicting user perspectives when exploring personalization in addition to considering the compute constraints.

\section{Conclusion}
This work addresses a critical gap in NLP concerning the personalization of LLMs. While LLMs have achieved remarkable performance in various tasks, their generalization capabilities have predominantly followed a "one-size-fits-all" paradigm. This approach falls short of meeting the diverse linguistic and communicative preferences of individual users. The \datasetname{} Benchmark introduced in this paper serves as an effort to evaluate the personalization capabilities of LLMs across a spectrum of tasks. \datasetname{}, presents a unique challenge by emphasizing scenarios where identical inputs necessitate diverse model outputs. The reported results showcase the inherent challenges posed by the \datasetname{} benchmark and advocate for the continued exploration of effective personalization strategies.

\section{Limitations}
The \datasetname{} Benchmark while designed to capture diverse user perspectives, may not fully represent the intricacies of all real-world communication scenarios. The dataset's construction involved a careful curation process, but the authors acknowledge that the complexities of individual preferences and linguistic nuances are vast and varied. In this work, user perspective is modeled solely based on the user's output preferences. Factors such as age, gender, and other potentially important demographics are not considered. 

In addition, the personalization methodologies explored in this study may not encompass the entire spectrum of potential approaches. The field of NLP is dynamic, and emerging techniques could offer alternative solutions to the challenges presented. Personalization in LLMs is an evolving research area, as such there may be relevant strategies released recently that were not covered in this work.

\bibliography{custom}

\appendix

\section{Additional Dataset Details}
\label{apd:datasets}

In this section, we detail the datasets in our \datasetname{} benchmark, including dataset construction, representative samples, and task instructions. 

\begin{table*}[th]
    \centering \footnotesize
    \begin{tabular}{ l l c m{0.44\textwidth} }
        \toprule
        \multicolumn{2}{c}{\thead{Dataset}} & \multirow{2}{*}[-0.25em]{\thead{\# Unique\\Texts}} & \multirow{2}{*}[-0.25em]{\thead{Labels}} \\
        \addlinespace[0.25em] \cline{1-2} \addlinespace[0.5em] 
        \thead{Domain} & \thead{Name} & \\

        \midrule
        \multirow{6}{*}[-3.25em]{Hate+Abuse} 
            & HateXplain & 20K  & [hateful, offensive, normal] \\
            \addlinespace[0.25em] \cline{2-4} \addlinespace[0.5em] 
            & GabHate & 28K & [Hateful, Non-hateful] \\
            \addlinespace[0.25em] \cline{2-4} \addlinespace[0.5em] 
            & \makecell[l]{Measuring\\HateSpeech} & 50K  & Hate speech scale: [0, 1, 2] \\
            \addlinespace[0.25em] \cline{2-4} \addlinespace[0.5em] 
            & TweetEval & 200 & [Hateful, Non-hateful] \\
            \addlinespace[0.25em] \cline{2-4} \addlinespace[0.5em] 
            & \makecell[l]{Unhealthy\\Conversations} & 44K & [healthy, unhealthy] \\
            \addlinespace[0.25em] \cline{2-4} \addlinespace[0.5em] 
            & \makecell[l]{WikiDetox\\Aggression} & 100K & [Aggressive, Normal] \\

        \midrule
        \multirow{5}{*}[-3em]{Sentiment} 
            & GoEmotion & 58K & [anger, disgust, fear, joy, sadness, surprise] \\
            \addlinespace[0.25em] \cline{2-4} \addlinespace[0.5em] 
            & StudEmo & 5K & [joy, trust, anticipation, surprise, fear, sadness, anger, disgust, valence, arousal] \\
            \addlinespace[0.25em] \cline{2-4} \addlinespace[0.5em] 
            & \makecell[l]{Subjective Discourse\\(response)} & \multirow{3}{*}[-1.25em]{1K}  & [answer+direct, answer+over-answer, shift+correct, shift+dodge, can't answer+honest, can't answer+lying] \\
            & \makecell[l]{Subjective Discourse\\(question sentiment)} &  & [very-negative, negative, somewhat-negative, neutral, somewhat-positive, positive, very-positive] \\
            & \makecell[l]{Subjective Discourse\\(response sentiment)} &  & [very-negative, negative, somewhat-negative, neutral, somewhat-positive, positive, very-positive] \\

        \midrule
        \multirow{2}{*}[-0.25em]{Humor} 
            & Cockamamie  & 120K & [humorous, not-humorous] \\
            \addlinespace[0.25em] \cline{2-4} \addlinespace[0.5em] 
            & EPIC & 3K & [Ironic, Non-ironic] \\
        \bottomrule
    \end{tabular}
    \caption{Additional details on the \datasetname{} Benchmark. } 
    \label{tab:dataset-additional-details}
\end{table*}

\subsection{Dataset Details \& Construction}
We include datasets in various domains, including: 

\begin{itemize}
    \itemsep0.125em 
    \item \textbf{HateXplain} \citep{mathew2022hatexplain} contains posts on social media. Each post is classified into 3 classes: hate, offensive, or normal. 
    The dataset additionally contained annotations for the hate speech target community and the rationales. 
    We consider the post texts and the classification labels only. 
    
    \item \textbf{GabHate} \citep{Kennedy2018TheGH} has 28K posts from the social media platform Gab. Each post is annotated using a hierarchical coding typology of hate-based rhetoric, with hierarchical labels indicating dehumanizing and violent speech, vulgarity and offensive language, and targeted groups. 
    We only consider the top-level binary classification on hate speech. 

    \item \textbf{MeasuringHateSpeech} \citep{sachdeva-etal-2022-measuring} contains 50K social media comments from various platforms. 
    They are labeled by a large number (11K) of Crowdsource workers on Amazon Mechanical Turk\footnote{\url{https://www.mturk.com}}. 
    Each comment is annotated with 10 ordinal labels: sentiment, disrespect, insult, attacking/defending, humiliation, inferior/superior status, dehumanization, violence, genocide, and a 3-valued hate speech benchmark label. 
    This dataset adjusts for annotators' perspectives by aggregating the labels via faceted Rasch measurement theory (RMT). 
    We use the comment text and the 3-valued hate speech label only. 

    \item \textbf{TweetEval} \citep{röttger2022contrasting} comprises 200 Twitter posts, each annotated by 20 annotator groups of 3. Annotators were given a short definition of hate speech only, which encourages the subjectivity of annotators. The labels are binary classifications of hatefulness. 

    \item \textbf{UnhealthyConversations} \citep{price-etal-2020-six} consists of 44K comments labeled by crowd-source workers as either ``healthy'' or ``unhealthy''. It also contains six potentially unhealthy sub-attributes: (1) hostile; (2) antagonistic, insulting, provocative, or trolling; (3) dismissive; (4) condescending or patronizing; (5) sarcastic; and/or (6) an unfair generalization. We only consider the top-level binary classification of healthy conversations. 

    \item \textbf{WikiDetox Aggression} \citep{wulczyn2017ex} is a large collection of 100K online comments to English Wikipedia where crowd-source annotators label whether each comment is a personal attack. 

    \item \textbf{GoEmotion} \citep{demszky2020goemotions} is a large annotated dataset of 58k English Reddit comments, labeled for 27 emotion categories or Neutral. We reduce the emotion categories into 6 coarse categories with Ekman-style grouping. 
    Each comment can have multiple emotion labels. We drop texts with no labels annotated and texts labeled as `neutral'. 

    \item \textbf{StudEmo} \citep{ngo-etal-2022-studemo} comprises 5K customer reviews annotated by 25 people for 10 emotion dimensions: eight emotion dimensions from Plutchik’s model plus valence and arousal. 
    Valence has an intensity range of $[-3, +3]$ whereas each remaining category has a range of $[0, 3]$. 
    We treat the problem as multi-class classification where we keep categories with intensity value $\geq 1$ as positive labels. 

    \item \textbf{Subjective Discourse} \citep{ferracane-etal-2021-answer} consists of 1,000 question-response pairs from 20 witness testimonials in U.S. congressional hearings. 
    The study collects subjective interpretations from the crowdsource workers about the conversations on three aspects: the question, the response, and an explanation. 
    In particular, the annotator provides subjective assessments of the conversation acts and communicative intents of the responses, forming 6 response labels. 
    Each annotator also rates their sentiment toward the politicians and the witness on a 7-point scale. 
    We leverage the response labels, and the sentiments toward the politicians and witnesses to form 3 dataset versions. 
    To construct the text part, we join (question speaker detail, question text, response speaker detail, response text) by newlines. 

    \item \textbf{Cockamamie} \citep{pmlr-v97-gultchin19a} includes 120K words and phrases from GNEWS. The words and phrases are annotated by crowd workers fluent in English on whether they are humorous. 1,500 words are further annotated on six binary humor features. 
    We leverage the words and the initial binary ``humorous'' labels only. 

    \item \textbf{EPIC} \citep{frenda-etal-2023-epic} is made of 3K short social media (post, reply) pairs collected from Twitter and Reddit across five regional varieties of English. Annotators from five different counties were asked to provide a binary label (either \emph{Irony} or \emph{not-Irony}) for the \emph{Reply} text given the context provided by \emph{Post}.
    We template the (Post, Reply) pairs by adding ``message'' and ``reply'' prefixes. 
\end{itemize}

We summarize additional details for each dataset in Table \ref{tab:dataset-additional-details}. 
We split each user's data into train/dev/test sets by 80/10/10 splits. 

\subsection{Representative Samples}
In this section, we show representative samples for each dataset
where different user perspectives result in different labels 
for the same text input. 

\noindent \textcolor{red}{Warning: Offensive content is inevitable for datasets in the Hate+Abuse domain. }

\subsubsection{Representative Sample for HateXplain}
\textbf{Text}. 
right definitely not going back to the fag hag thing it was fun for a bit in my <number> but outrageously gay people are just too high strung for me it back to that reason i can not be a lesbian 

\noindent\textbf{Labels}. 


- normal

- hatespeech

- offensive

\subsubsection{Representative Sample for MeasuringHateSpeech}
\textbf{Text}. 
@L1LTR4P fucking losers wetbacks \#SorryNotSorry

\noindent\textbf{Labels}. 

- 0 

- 1 

- 2

\subsubsection{Representative Sample for GabHate}
\textbf{Text}. 
\#Breaking: Claiming bias against conservatives, @FreedomWatchUSA has filed a class action lawsuit against Google, Facebook, Instagram, Twitter, and Apple.

\noindent\textbf{Labels}. 


- Non-hateful

- Hateful

\subsubsection{Representative Sample for TweetEval}
\textbf{Text}. 
[USER] fuck Brett Farve redneck ass, he stuckup he don't give a damn lol he be on campus acting like he the shit

\noindent\textbf{Labels}. 

- Hateful

- Non-hateful

\subsubsection{Representative Sample for Unhealthy Conversations}
\textbf{Text}. 
They are poor because they are reliant on the drug trade and reliant on the drug trade because they are then poor. That cycle can be broken.

\noindent\textbf{Labels}. 

- healthy 

- unhealthy

\subsubsection{Representative Sample for WikiDetox Aggression}
\textbf{Text}. 
== Dougbiznatch again! ==

~\par

\noindent Hey I'm back. Gonna vandalize all day and no one can stop me! As you can tell I can't be stopped by banning. I'll be seeing alo tof you and the rest of the blacklisted admins for the next couple of weeks =P  

\noindent Dougbiznatch

\noindent\textbf{Labels}. 

- Aggressive 

- Normal

\subsubsection{Representative Sample for GoEmotion}
\textbf{Text}. 
Is he dead now from a tragic drowning accident? Asking for a friend.

\noindent\textbf{Labels}. 

- [sadness]

- [surprise]

\subsubsection{Representative Sample for StudEmo}
\textbf{Text}. 
We got to the Cycle On Hostel by chance in the middle of the night. There wasn't a single place available in other places. . . ...and it's very good that we didn't get to another place! First of all, great service: people who are open to others, nice and smiling, who help us with every time of day and night. Spacious hostel, rooms and bathrooms are clean. And besides all that - the location - nothing to add, nothing to take away. Literally 5 minutes away from Neptune and at the same time the building is situated in such a place that at night it is quiet despite such a short distance from the busiest street where it is full of tourists and children. If we will still have a chance to spend the night in Gdansk, we will surely come to Cycle On again. With a clear conscience I recommend

\noindent\textbf{Labels}. 

- [trust, anticipation, valence, arousal]

- [joy, trust, valence, arousal]

\subsubsection{Representative Sample for Subjective Discourse (response)}
\textbf{Text}. 
\noindent politician: JIM JORDAN, Ohio 

~\par 

\noindent Okay. Well, this is put out by the Exempt Organizations Division, same division where all these problems took place over the last three years. It came out, again, just five days after the comment period on the proposed rule ended at the end of February, and I want to just highlight a few of the questions that are asked. So there is a category that says what if the IRS needs more information about your (c)(4) application? New sample questions. So if we could put them up side-by-side.    Now, the first slide are the targeted questions that TIGTA said were inappropriate and that you agree are inappropriate, Judith Kendall agreed are inappropriate. Those are those questions. And now, just five days after the proposed rule comment period ends, you issue a newsletter from the Exempt Organizations Division highlighting the new questions you are going to ask, and I just want to look how similar the two questions are.    Let's just take the second category, whether an officer or director, etcetera, has run or will run for public office. The new question says this: Do you support a candidate for public office who is one of your founders, officers, or board members? It is basically the same. This reminds me of when I was in grade school and the teachers told us you shouldn't plagiarize, so you change a few words and basically plagiarize. This is the same thing.    So here is what I don't understand. If you are trying to comply with the TIGTA report, if the new (c)(4) rule is a way to deal with what the audit said and not as what I believe is a continuation of the project Lois Lerner started, why are you asking the same darn questions? 

~\par 

\noindent witness:  The Hon. John Koskinen, Commissioner, Internal Revenue Service 

~\par 

\noindent As I noted, I haven't seen that and can't read it on the chart. I would be delighted to sit down and go over all of those questions with you and with the exempt organizations. All of the TIGTA report didn't blanket say you should never ask questions about this. Thank you for the chart.

\noindent\textbf{Labels}. 

- can't answer+lying

- can't answer+honest

- shift+dodge

\subsubsection{Representative Sample for Subjective Discourse (question sentiment)}
\textbf{Text}. 
politician: RANDY NEUGEBAUER, Texas 

~\par

\noindent And, as you're aware, Section 972 of Dodd-Frank requires an issuer of securities to disclose the annual proxy statement, the reason why the issuer has chosen to allow the same person to serve as the board chairman and the CEO. This year, Wells states that your dual role is a result of your extensive experience and knowledge regarding the company and provides the most efficient leadership of the board and the company.    Mr. Stumpf, do you think it's a good idea for the same person to serve as both chairman of the board and CEO? 

~\par

\noindent witness:  Stumpf, John G., Chairman and Chief Executive Officer, Wells Fargo and Company

~\par

\noindent Thank you, Congressman, for that question.    In our company, we have 14 outside directors. We have a lead director. All directors are New York Stock Exchange-independent, by their standards. I am not a member of any standing committee of that board. The independent directors and the lead director help set the agenda for the boards. They always have meetings that are in executive session without me.    And as you probably read about what happened this weekend--because we filed an 8-K yesterday about actions that they took as an independent board, and I was not part of that. So the board acts quite independently.

\noindent\textbf{Labels}. 

- negative

- neutral

- somewhat-negative

- somewhat-positive

\subsubsection{Representative Sample for Subjective Discourse (response sentiment)}
\textbf{Text}. \noindent politician: JERROLD NADLER, New York, Chairman 

~\par 

\noindent It is our understanding that at least one briefing occurred between December--between your decision not to recuse yourself on December 19th and 6 days later, Christmas day. Is that correct? Simple enough question, yes or no.

~\par 

\noindent witness:  Matthew G. Whitaker, Acting Attorney General, U.S. Department of Justice Oral Testimony 

~\par 

\noindent Mr. Chairman, I, again, what is the basis for your question? You are saying that it is your understanding----

\noindent\textbf{Labels}. 

- negative

- neutral

- somewhat-negative

- somewhat-positive

\subsubsection{Representative Sample for Cockamamie}
\textbf{Text}. 
judokas

\noindent\textbf{Labels}. 

- humorous

- not-humorous

\subsubsection{Representative Sample for EPIC}
\textbf{Text}. 
message: ``Because the last generation was Gen. Z and there's no where to go past that, so we gotta start from the beginning.''

\noindent reply: ``but we should have just named the first generation ``Alpha'' instead of doing it now''

\noindent\textbf{Labels}. 

- Ironic 

- Non-ironic

\subsection{Task Instructions}
In this section, we show the task instructions prepended to prompts in training and inference for each dataset.

\subsubsection{Instruction for HateXplain}

\noindent Please review the following text and indicate if it has the presence of hate speech or is offensive in nature. Respond `hatespeech' if the text contains hate speech, `offensive' if the text does not contain hate speech but is offensive and `normal' if the text is neither offensive nor contains hate speech.


\subsubsection{Instruction for MeasuringHateSpeech}

\noindent On a scale of 0-2 rate the level of hatespeech contained in the following text. Respond with `0' if the text does not contain hate speech, `1' if the text contains some hate speech, and `2' if the text contains a lot of hate speech.

\subsubsection{Instruction for GabHate}

\noindent Please review the following text and indicate if it has the presence of hate speech. Respond `Hateful' if the text contains hate speech and `Non-hateful' if the text does not contain hate speech.

\subsubsection{Instruction for TweetEval}

\noindent Please review the following text and indicate if it has the presence of hate speech. Respond `Hateful' if the text contains hate speech and `Non-hateful' if the text does not contain hate speech.

\subsubsection{Instruction for Unhealthy Conversations}

Please review the following text and indicated if it is `healthy' or `unhealthy'. Respond `healthy' if the text is healthy and `unhealthy' if the text can be considered hostile, antagonistic, condescending, dismissive or an unfair generalization.

\subsubsection{Instruction for WikiDetox Aggression}

Please review the following text and indicate if it has the presence of malicious remark to a person or group. Respond `Aggressive' if the text contains a personal attack and `Normal' if the text does not contain a personal attack.

\subsubsection{Instruction for GoEmotion}

Please analyze the following text and assign one or more appropriate emotion labels. Emotion labels include happiness, sadness, anger, surprise, joy, fear, disgust. You can select one or multiple emotion labels that best capture the emotional content of the text. Respond with the emotion labels separated by a comma.

\subsubsection{Instruction for StudEmo}

Please analyze the following text and assign one or more appropriate emotion labels. Emotion labels include joy, trust, anticipation, surprise, fear, sadness, disgust, anger, valence, and arousal. You can select one or multiple emotion labels that best capture the emotional content of the text. Respond with the emotion labels separated by a comma.

\subsubsection{Instruction for Subjective Discourse (response)}

Please analyze the following text and indicate how the witness responded to the question. Respond with `answer' if they answered the question reasonably, `cant-answer-lying' if they could not answer and are lying, `cant-answer-sincere' if they could not answer but are honest about it, `shift-dodge' if they shifted the topic with the intent of dodging the question, `answer\_overans-sway' if they over answered the question with the intention of swaying or `shift-correct' if they shifted the topic with the intention of clarifying the question.

\subsubsection{Instruction for Subjective Discourse (question sentiment)}

Please analyze the following text and rate your sentiment towards the questioners. Sentiment labels include `somewhat-positive', `positive', `very-positive', `somewhat-negative', `very-negative', `neutral' and `negative'. Respond with the sentiment label that best captures your sentiment towards the questioners.

\subsubsection{Instruction for Subjective Discourse (response sentiment)}

Please analyze the following text and rate your sentiment towards the witness. Sentiment labels include `somewhat-positive', `positive', `very-positive', `somewhat-negative', `very-negative', `neutral' and `negative'. Respond with the sentiment label that best captures your sentiment towards the witness.

\subsubsection{Instruction for Cockamamie}

Please rate whether the following text is funny or not funny. Respond `yes' if you think the text is funny and `no' if you think the text is not funny.

\subsubsection{Instruction for EPIC}

Irony is a figurative language device that conveys that opposite of literal meaning, profiling intentionally a secondary or extended meaning. Please review the following message and reply and indicate if it has the presence of irony. Respond `Ironic' if the reply if you think the reply is ironic and `Non-ironic' if you think the reply is not ironic.

\end{document}